\documentclass[10pt,twocolumn,letterpaper]{article}

\usepackage{cvpr}      

\usepackage{graphicx}
\usepackage{amsmath}
\usepackage{amssymb}
\usepackage{booktabs}
\usepackage[noend]{algpseudocode}

\usepackage{algorithmicx,algorithm}

\usepackage[pagebackref,breaklinks,colorlinks]{hyperref}

\usepackage[capitalize]{cleveref}
\crefname{section}{Sec.}{Secs.}
\Crefname{section}{Section}{Sections}
\Crefname{table}{Table}{Tables}
\crefname{table}{Tab.}{Tabs.}

\usepackage{color}
\definecolor{runpei-blue}{RGB}{0, 113, 188}

\begin{document}

\title{SimpleBEV: Improved LiDAR-Camera Fusion Architecture for 3D Object Detection}

\author{
    Yun Zhao,~
    Zhan Gong,~
    Peiru Zheng,~
    Hong Zhu,~
    Shaohua Wu\\
    IEIT\\
    {\tt\small\{zhaoyun02,gongzhan01,zhengpeiru,zhuhongbj,wushaohua\}@ieisystem.com}
}

\maketitle
\begin{abstract}
  More and more research works fuse the LiDAR and camera information to improve the 3D object detection of the autonomous driving system. Recently, a simple yet effective fusion framework has achieved an excellent detection performance, fusing the LiDAR and camera features in a unified bird's-eye-view (BEV) space. In this paper, we propose a LiDAR-camera fusion framework, named SimpleBEV, for accurate 3D object detection, which follows the BEV-based fusion framework and improves the camera and LiDAR encoders, respectively. Specifically, we perform the camera-based depth estimation using a cascade network and rectify the depth results with the depth information derived from the LiDAR points. Meanwhile, an auxiliary branch that implements the 3D object detection using only the camera-BEV features is introduced to exploit the camera information during the training phase. Besides, we improve the LiDAR feature extractor by fusing the multi-scaled sparse convolutional features. Experimental results demonstrate the effectiveness of our proposed method. Our method achieves 77.6\% NDS accuracy on the nuScenes dataset, showcasing superior performance in the 3D object detection track.
\end{abstract}

\section{Introduction}
\label{sec:intro}
3D object detection plays an indispensable role in an autonomous driving perception system, which recognizes and localizes the object in the 3D traffic environment~\cite{Sun2019ScalabilityIP,Caesar2019nuScenesAM,Xu2021FusionPaintingMF,bevformer,Chen2016Multiview3O}. Multiple sensors have been equipped on self-driving vehicles to obtain sufficient and accurate perception results. The camera and LiDAR sensors are extensively investigated among all the onboard sensors. The LiDAR points~\cite{2017PointNetplusplus,pointpillars,second} provide precise localization and geometry information, and the cameras~\cite{bevdet,bevformer,fcos3d} offer abundant semantic information. As these two kinds of sensors provide complementary characteristics, a lot of research works~\cite{Liang2018DeepCF,Vora2019PointPaintingSF,Liang2022BEVFusionAS,Bai2022TransFusionRL,Yang2022DeepInteraction3O} fuse the LiDAR and camera data to enhance the 3D object detection performance.

Recently, the BEV-based perception method has received considerable attention as it is an intuitive representation of driving scenarios~\cite{Chen2016Multiview3O,2022Vision} and is fusion-friendly for multi-view cameras~\cite{bevformer,bevdet,bevdet4d} and different kinds of sensors~\cite{Liu2022BEVFusionMM,Bai2022TransFusionRL,Liang2022BEVFusionAS}. A series of methods~\cite{Li2022DeepFusionLD,Bai2022TransFusionRL,Yan2023CrossMT} utilize a transformer-based architecture to fuse the LiDAR and camera information by performing the cross-attention on the LiDAR features and image features. Differently, some works~\cite{Liang2022BEVFusionAS,Liu2022BEVFusionMM} implement the LiDAR-camera fusion based on the aligned BEV feature maps. Though simple, the BEV-based fusion framework achieves an excellent detection performance. In this paper, we build a LiDAR-camera fusion framework based on the BEVFusion~\cite{Liang2022BEVFusionAS} by further exploiting the camera information and improving the LiDAR feature extractor.

For exploiting the camera information, we enhance the depth estimation module and bring an auxiliary detection branch. The depth estimation module plays a crucial role in camera-based 3D object detection. A precise depth result benefits the feature alignment when fusing the LiDAR and camera BEV feature maps. So, we introduce a two-stage cascade network for better image-based depth estimation and rectify the estimated depth map with the depth information derived from the LiDAR points. The LiDAR modality plays a prominent role against the camera modality when integrating information from LiDAR and camera data. To further exploit the camera information during jointly training the whole model, we introduce an auxiliary branch that utilizes only the camera-BEV feature to implement 3D object detection.

Besides, we improve the LiDAR feature extractor by fusing the multi-scaled sparse convolutional features. To reduce the consumption of computation and memory cost, the 3D voxel features are first encoded into the BEV space. Then, these multi-scaled LiDAR-BEV feature maps are fused to generate an expressive BEV feature map.

Experimental results demonstrate that the introduced auxiliary branch and the improved camera/LiDAR feature extractor can effectively increase 3D object detection performance. Moreover, with the model ensemble and test-time augmentation, our model achieves the best NDS score at the nuScenes leaderboard.

Our contributions are summarized as follow:
\begin{itemize}
  \item We build a multi-modal detection model for 3D object detection. It follows the framework of the BEVFusion~\cite{Liang2022BEVFusionAS}, but brings an auxiliary branch to exploiting the camera information during the training phase. Moreover, we improve the camera-based depth estimator and the LiDAR-based feature encoder to provide more effective features for multi-modality fusion.
  \item The proposed method, SimpleBEV, achieves state-of-the-art 3D object detection performance on the nuScenes dataset.
\end{itemize}


\section{Related Works}
\noindent \textbf{Camera-based 3D object detection.} Early works~\cite{fcos3d,2020RTM3D} are proposed for monocular 3D object detection. Generally, they implement the 2D object detection based on the image and then project the 2D results into the 3D space using a second stage. However, this intuitive detection strategy suffers from the elaborate post-processes to achieve robust results when dealing with the inputs from surrounding cameras. Recently, vision BEV perception methods~\cite{lss,bevdet,bevformer} received enormous attention in industry and academia. These architectures transform the features from multiple images into a unified BEV frame~\cite{pointpillars}. The BEV features can be directly fed to many downstream tasks and are fusion-friendly. These methods can be broadly classified into two categories on the basis of the transformation mode~\cite{2022Vision}: ``geometry-based transformation'' and ``network-based transformation''. The representative ``geometry-based'' methods \cite{lss,bevdet,bevdepth} adopt explicit depth estimation and project the extracted features into the 3D space based on the physical principles. \cite{bevdepth} applies the LiDAR data to supervise the depth prediction training and \cite{bevdet4d} introduces temporal cues to increase the 3D object detection performance. Differently, the ``network-based'' methods use the neural network to map the image features to the BEV space implicitly. A lot of works~\cite{bevformer,jiang2022polar,2022PETR} use transformers to translate the image features to the BEV space. They all use the deformable transformer~\cite{2021Deformable} to reduce computation and memory costs.

\noindent \textbf{LiDAR-based 3D object detection.} The mainstream 3D object detection methods can be categorized into point-based~\cite{2017PointNet,2017PointNetplusplus} and voxel-based~\cite{second,focalsconv-chen,2021SE} methods. The point-based methods~\cite{2017PointNet,2017PointNetplusplus} directly operate the irregular LiDAR points and exploit the spatial information. Differently, the voxel-based methods~\cite{second,focalsconv-chen,2021SE} first transform the unordered LiDAR points into the volumetric mode with a pre-defined grid size and then apply 2D/3D CNNs on the regular voxels to obtain the detection results. Recently, some methods~\cite{2020PV,2021HVPR} have integrated 3D voxel networks and point-based networks to achieve more representative features.

\noindent \textbf{Multi-modal 3D object detection.} The LiDAR and camera information are complementary. The LiDAR points can provide precise spatial information for object location, and the images give abundant contextual information for object classification. To obtain accurate surrounding information for autonomous vehicles, many researchers endeavor to effectively fuse the information from cameras and LiDAR for accurate 3D object detection. According to the fusion operation, the camera-LiDAR fusion methods can be divided into three categories~\cite{Mao20223DOD}: ``early-fusion'', ``intermediate-fusion'' and ``late-fusion''. The ``early-fusion'' methods mainly first implement image information (features~\cite{Wang2021PointAugmentingCA}, semantic labels~\cite{Vora2019PointPaintingSF}, or bounding boxes~\cite{Qi2017FrustumPF}) and feed the results to the LiDAR-based branch to achieve the final detection. These methods need an additional complicated 2D network and suffer from the detection of the object with few LiDAR points. The ``late-fusion'' method~\cite{Pang2020CLOCsCO} fuses the results from the independent camera and LiDAR branches. Despite its efficiency, this method limits the exploitation of rich and complementary information from different modalities. The ``intermediate-fusion'' methods gain the most attention in industry and academia. Early research works~\cite{Chen2016Multiview3O,Ku2017Joint3P} generate 3D object proposals based on LiDAR or LiDAR-camera information and fuse the LiDAR and camera features extracted based on the object proposals. In recent years, many BEV-related fusion methods~\cite{Liang2022BEVFusionAS,Liu2022BEVFusionMM,Li2022DeepFusionLD,Bai2022TransFusionRL} have been proposed inspired by the vision BEV representation. \cite{Liang2022BEVFusionAS,Liu2022BEVFusionMM} extract camera BEV features using LSS~\cite{lss} and fuse them with the LiDAR BEV features. \cite{Li2022DeepFusionLD} uses the LiDAR features as the queries to fuse the image and LiDAR features. \cite{Bai2022TransFusionRL} builds a two-stage pipeline where the first stage produces initial 3D bounding boxes and the second stage associates and fuses the object queries with the image features for better detection results. \cite{Yan2023CrossMT} treats the image and LiDAR features as tokens and directly implements 3D object detection using transformers. To further exploit the camera information during fusion, \cite{Yang2022DeepInteraction3O} applies two individual branches to perform representational interaction and sequential modules for predictive interaction. Our method builds on the BEVFusion~\cite{Liang2022BEVFusionAS} method and strengthens the camera and LiDAR branches to achieve better 3D object detection performance.

\section{Method}
\begin{figure*}[t]
    \centering
    \includegraphics[width=\textwidth]{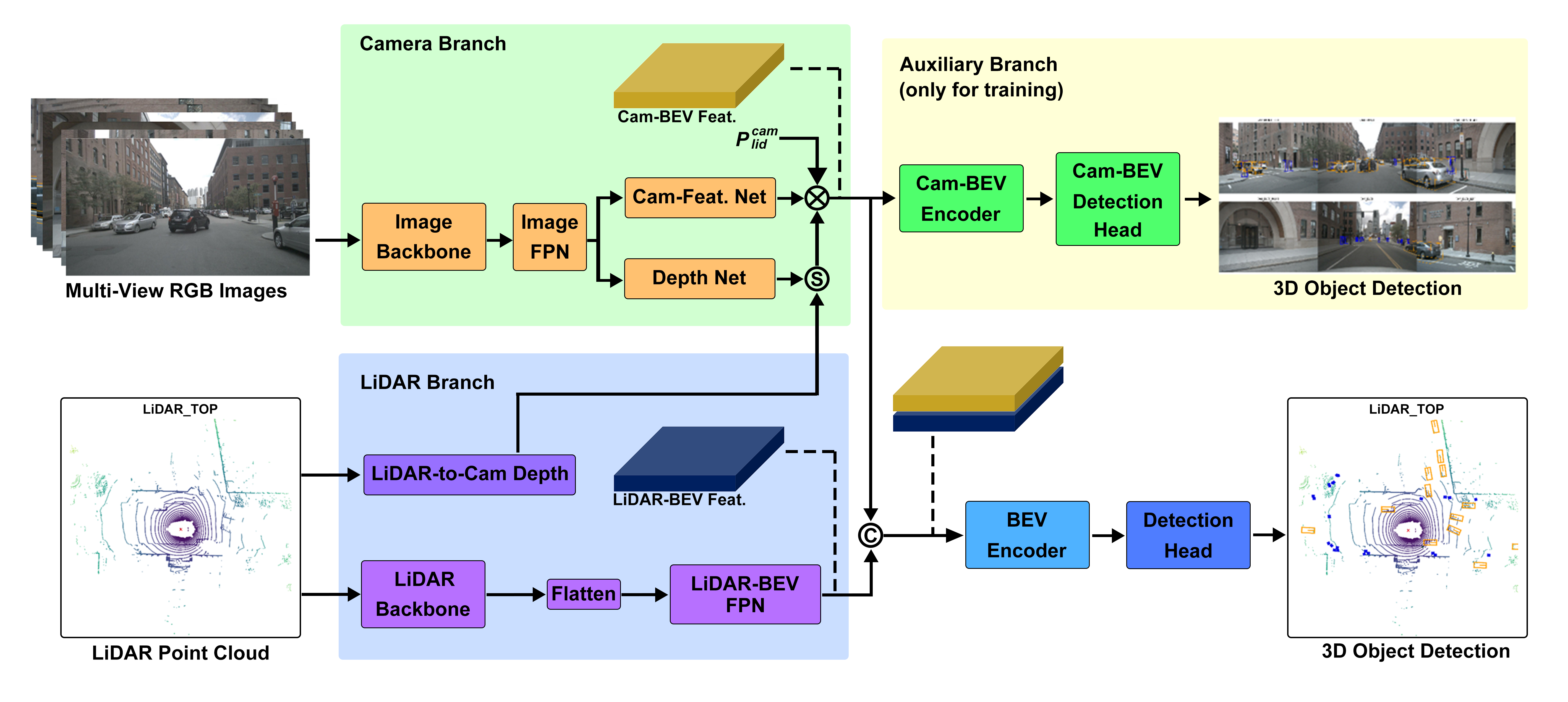}
    \caption{An overview of SimpleBEV framework. Two branches separately extract the features based on the LiDAR points and multi-view images. The features are transformed into a unified BEV space. The auxiliary branch only works in the training phase.}
    \label{fig:framework}
\end{figure*}

We design a multi-modal 3D object detector, SimpleBEV, based on the camera and LiDAR data, whose framework is shown in Fig.~\ref{fig:framework}. We first introduce the camera-related branches and the LiDAR branch. The camera-related branches consist of a camera branch to extract the image features and project them into the BEV space, and an auxiliary branch to better exploit the information from cameras during the training phase. Then, we present the BEV encoder and the detection head for the final detection task.

\subsection{Camera related branches}
\label{sec:camera_related_branches}

\begin{figure}
  \centering
  \includegraphics[width=\columnwidth]{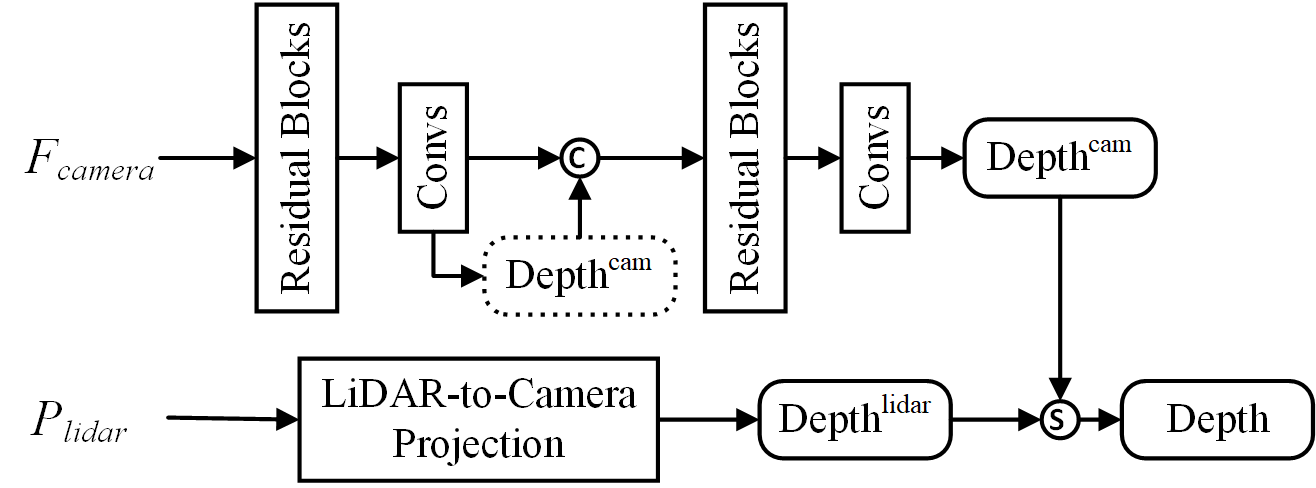}
  \caption{Pipeline of depth estimation.}
  \label{fig:depth_estimation}
\end{figure}

\noindent\textbf{Camera branch.}
The input multi-view images are first encoded into deep features through a shared image encoder which consists of an image backbone for feature extraction and a simple FPN neck to fuse multi-scale features. Specifically, the ConvXt-Tiny~\cite{liu2022convnet} is adopted as the image backbone to extract representative image features. The feature maps from different stages of the image backbone are fed to the FPN neck to exploit the scale variate representation. Then, the feature maps from the specified layer are used to generate the camera-BEV feature map.

Given the $i$-th image feature map $\textbf{F}^{I_i}_{cam} \in R^{H \times W \times C_{\text{im}}}$, we transform the image features into the BEV space following a similar pipeline in LSS~\cite{lss}. The image features are first used to estimate the pixel-wise depth distributions $\textbf{D}_{cam}^{I_i} \in R^{H \times W \times D}$, where $D$ denotes the number of discretized depth bins. Then, each image feature is weighted by the probabilities of different depth bins and projected to the 3D coordinate to form the frustum features. The 3D features from multiple cameras are all transformed into the LiDAR coordinate and form the camera-BEV feature map $\textbf{F}^B_{cam} \in R^{X\times Y \times C_{\text{cam}}}$ through voxelization and sum pooling along the height. $X$ and $Y$ represent the grid size along the $x$-axis and $y$-axis of the BEV coordinate, respectively.

The depth estimation in the above feature transformation procedure plays a critical role in camera-based 3D object detection. A better depth predictor benefits the alignment of the camera-BEV and LiDAR-BEV features. To improve the precision of the depth estimation, we modify the depth estimation network and introduce the LiDAR data to generate the precise depth. The pipeline is shown in Fig.~\ref{fig:depth_estimation}. A two-stage cascade structure is built to gain the camera-based depth map $\textbf{D}_{cam}^{I_i}$. The output depth map from the first stage is concatenated with the feature map from the first stage, and the fused feature map is fed into the second stage. Meanwhile, the LiDAR points are transformed into the $i$-th camera coordinate and projected into the image coordinate to form the depth map $\textbf{D}_{lid}^{I_i} \in R^{H \times W \times D}$. Considering the projected points on the feature map are sparse, we introduce a mask map $\textbf{M}_{lid}^{I_i} \in \{0, 1\}^{H \times W}$ to represent whether the pixel of the feature map is labeled by the LiDAR points (denoted as $1$) or not (denoted as $0$). The depth of the pixel $(u, v)$ in the final depth map $\textbf{D}^{I_i} \in R^{H \times W \times D}$ is calculated as $\textbf{D}^{I_i}(u, v) = \textbf{D}_{lid}^{I_i}(u, v) \cdot \textbf{M}_{lid}^{I_i}(u,v) + \textbf{D}_{cam}^{I_i}(u, v) \cdot (1 - \textbf{M}_{lid}^{I_i}(u,v))$. In other words, the final depth map is generated by filling the holes on the sparse LiDAR-based depth map with the estimated depth map based on the image features. The fused depth map is used for the image feature projection.

\noindent\textbf{Auxiliary branch.}
An auxiliary branch is introduced to exploit further the camera information, which is activated during the training phase. A camera-BEV encoder encodes the BEV feature from the camera branch. An anchor-free-based detection head is introduced to implement the 3D object detection task. The camera-BEV encoder consists of multi-layer convolutions and multi-scale feature fusion modules. The auxiliary detection head follows the structure of CenterHead~\cite{Yin2020Centerbased3O} to perform 3D object detection using only the camera-BEV features.

\subsection{LiDAR branch}
\label{sec:lidar_branch}
\begin{figure}
  \centering
  \includegraphics[width=\columnwidth]{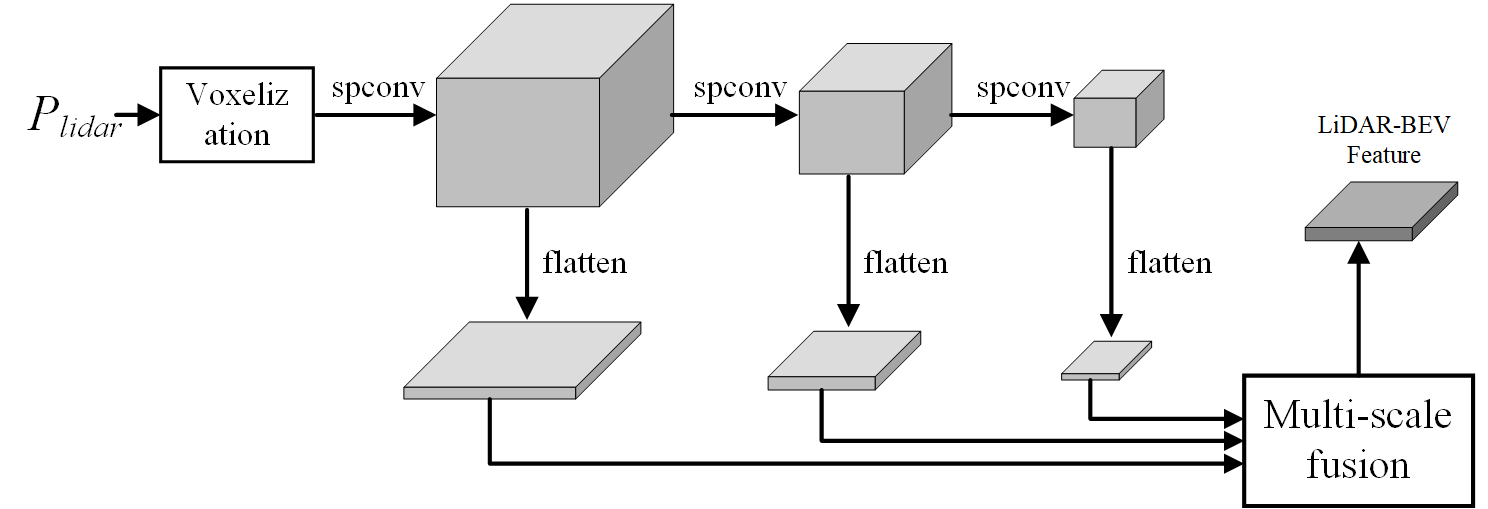}
  \caption{Overview of the LiDAR branch structure. $P_{lidar}$ represents the LiDAR points.}
  \label{fig:lidar_branch}
\end{figure}

The LiDAR branch follows a similar pipeline as SECOND~\cite{second} to extract 3D features and fuses multi-scaled features from different stages. The framework is shown in Fig.~\ref{fig:lidar_branch}. The raw points are first converted to the voxel features. Then, multiple sparse 3D convolution layers are sequentially applied to the features to generate multi-scale 3D features. To enhance the capability of the LiDAR-based features to capture multi-scale objects, we introduce the multi-scale feature fusion strategy. The multi-scaled 3D features from different stages are first transformed into multiple 2D BEV features. We apply multiple 3D convolutions to compress the $z$-dimension and concatenate the features along the $z$-dimension to transform the 3D features into the 2D BEV features. Then, multiple up-sampling and convolution operations are utilized to fuse the multiple BEV feature maps. The final LiDAR-BEV features $\textbf{F}^B_{\text{lid}}\in R^{X\times Y \times C_{\text{lid}}}$ are fed to the camera-LiDAR feature fusion module.

\subsection{BEV encoder \& Detection head}
The fused BEV features $\textbf{F}^B_{\text{fuse}}\in R^{X\times Y \times C_{\text{fuse}}}$ are generated by concatenating the camera-BEV feature $\textbf{F}^B_{\text{cam}}$  and the LiDAR-BEV feature $\textbf{F}^B_{\text{lid}}$. Then, the fused features are further encoded in the BEV space. The BEV encoder enhances the BEV features by multiple convolutions and combines the multi-scale features.

We adopted the mature transformer-based head~\cite{Bai2022TransFusionRL} and the center heatmap head~\cite{Yin2020Centerbased3O} for the final detection task and the auxiliary detection task, respectively. Specifically, the transformer-based head uses the fused BEV features, and the center heatmap head utilizes the camera BEV features.

\subsection{Training}
The model is trained by minimizing the sum of the following losses:
\begin{equation}
  \label{eq:loss}
  L=L_{fusion} + L_{aux} + L_{depth}.
\end{equation}
Here, $L_{fusion}$ represents the detection loss based on the fusion BEV features. We use the identical loss function as \cite{Bai2022TransFusionRL}. $L_{aux}$ denotes the detection loss from the auxiliary branch using only the camera-BEV features. $L_{depth}$ is the depth loss to train the depth net in the camera branch. The ground truth depth derives from the LiDAR data.

\section{Experiments}

\subsection{Experimental setup}

\noindent \textbf{Dataset.}
The experiments are conducted on the nuScenes dataset~\cite{Caesar2019nuScenesAM}. It provides the point cloud information using $1\times$ 32-beam LiDAR with 20HZ capture frequency and the image information from $6\times$ surrounding cameras with 12HZ capture frequency. The provided images from each camera are of the same resolution: 1600 $\times$ 900. This dataset consists of 1000 scenes with annotated 3D bounding boxes, which are divided into train/validation/test subsets with 700/150/150 scenes, respectively. We evaluate the models using the metrics of the mean average precision (mAP) and the nuScenes detection score (NDS) on 10 object detection results.

\noindent \textbf{Implementation details.}
We implement our method based on the code base mmdetection3d~\cite{mmdet3d2020}. During the evaluation, the input images are scaled and cropped to 256$\times$704 resolution, and the camera branch generates the features with $1/8$ input resolution. We set the voxel size to (0.075m, 0.075m, 0.2m) for X,Y,Z axis, respectively. The detection range is set to [-54m, 54m] for X and Y axis and [-5m, -3m] for Z axis. The BEV grid size is set to 0.6m.

We train the model using AdamW~\cite{Loshchilov2017DecoupledWD} optimizer with a one-cycle learning rate strategy~\cite{2017Super}. The model is trained for 20 epochs, in which the GT-Pasted data augmentation strategy is stopped after the $15$th epoch. No CBGS~\cite{zhu2019class} is used. During testing, we enlarge the image resolution to 640$\times$1600. For the online submission, we adopt the test-time augmentation (TTA) with multiple yaw rotations and global scales. Meanwhile, we train multiple models with additional voxel size (0.05m, 0.05m, 0.2m) and BEV grid sizes between 0.3m and 0.6m. The results are fused by the weighted box fusion (WBF) strategy~\cite{Solovyev2019WeightedBF}.

\subsection{Comparison to the state-of-the-art methods}

\begin{table*}[t]
	\begin{center}
	\scalebox{0.925}{	
  \begin{tabular}{l|c|cc|cccccccccc}
    \toprule
    Method  & Mod.  & mAP   & NDS           &  Car & Truck & C.V. & Bus & T.L. & B.R. & M.T. & Bicycle & Ped. & T.C.\\

    \midrule

    CenterPoint~\cite{Yin2020Centerbased3O} & L & 60.3 & 67.3 & 85.2 & 53.5 & 20.0 & 63.6 & 56.0 & 71.1 & 59.5 & 30.7 & 84.6 & 78.4 \\
    Focals Conv~\cite{focalsconv-chen} & L & {63.8} & {70.0} & 86.7 & 56.3 & {23.8} & 67.7 & 59.5 & {74.1} & 64.5 & 36.3 & {87.5} & {81.4} \\
    TransFusion-L~\cite{Bai2022TransFusionRL} & L & {65.5} & {70.2} & {86.2} & {56.7} & {28.2} & {66.3} & {58.8} & {78.2} & {68.3} & {44.2} & {86.1} & {82.0} \\

    \midrule
    PointAugmenting~\cite{Wang2021PointAugmentingCA}\dag & LC  & 66.8 & 71.0 & {87.5} & 57.3 & 28.0 & 65.2 & 60.7 & 72.6 & 74.3 & 50.9 & 87.9 & 83.6 \\
    TransFusion~\cite{Bai2022TransFusionRL} & LC & {68.9} & {71.7} & 87.1 & 60.0 & {33.1} & 68.3 & 60.8 & {78.1} & 73.6 & 52.9 & {88.4} & {86.7} \\
    BEVFusion~\cite{Liang2022BEVFusionAS} & LC & {69.2} & {71.8} & 88.1 & 60.9 & 34.4  & 69.3 & 62.1 & 78.2 & 72.2 & 52.2 & 89.2 & 85.5 \\
    BEVFusion~\cite{Liu2022BEVFusionMM} & LC & {70.2} & {72.9} & {88.6} & 60.1 & {39.3} & 69.8 & {63.8} & 80.0 & 74.1 & 51.0 & 89.2 & 86.5 \\
    CMT~\cite{Yan2023CrossMT}           & LC & 70.4     & 73.0  & 87.2  & 61.5 & 37.5   & 72.4 & 62.8   & 74.7 & 79.4 & 58.3 & 86.9 & 83.2 \\
    DeepInteraction~\cite{Yang2022DeepInteraction3O}      & LC & {70.8} & {73.4} & 87.9 & {60.2} & 37.5 & {70.8} & {63.8} & {80.4} & {75.4} & {54.5} & {91.7} & {87.2} \\
    Focals Conv-F~\cite{focalsconv-chen}\dag & LC & {70.1} & {73.6} & 87.5 & 60.0 & 32.6  & 69.9 & 64.0 & 71.8 & 81.1 & 59.2 & {89.0} & {85.5} \\
    BEVFusion~\cite{Liu2022BEVFusionMM}\S & LC & 75.0   & 76.1   & {90.5} & {65.8} & 42.6 & {74.2} & {67.4} & 81.1 & 84.4 & 62.9 & 91.8 & 89.4 \\
    DeepInteraction~\cite{Yang2022DeepInteraction3O}\S & LC                   & {75.6} & {76.3} & 88.3   & 64.3 & {44.7} & {74.2} & 66.0 & {83.5} & {85.4} & {66.4} & {92.5} & {90.9} \\
    \textbf{SimpleBEV}(ours)\S & LC                 & 75.7   & 77.6   & 89.0     & 69.2 &  44.4  & 75.2 & 70.5 & 78.4   & 81.2  & 69.1   & 91.7 & 87.9   \\
    \bottomrule
  \end{tabular}
  }
\caption{Comparison on the nuScenes test set. `C.V.', `T.L.', `B.R.', `M.T', `Ped.' and `T.C.' are short for construction vehicle, trailer, barrier, motor, pedestrian and traffic cone, respectively. `L' and `C` are short for LiDAR and camera.
    \dag~denotes TTA is applied, and \S~denotes that TTA and model ensemble are both used.}
\label{table:comparison}
	\end{center}
\end{table*}

We compare our methods with the state-of-the-art methods on the nuScenes test dataset. The results are shown in Tab.~\ref{table:comparison}. Our model achieves the best results on the mAP and NDS metrics. Meanwhile, we also compare our results with the state-of-the-art methods on the nuScenes validation set. As shown in Tab.~\ref{table:val_comparison}, our method surpasses the baseline of our method, BEVFusion~\cite{Liang2022BEVFusionAS}, by 3.5\% mAP and 2.5\% NDS. The excellent performance on both mAP and NDS demonstrates the effectiveness of our method.

\begin{table}[t]
	\begin{center}
	\scalebox{0.95}{	
  \begin{tabular}{l|c|cc}
    \toprule
    Method      & Modality  & mAP & NDS\\
    \midrule
    BEVDet4D~\cite{bevdet4d}    & C     & 42.1          & 54.5          \\
    BEVFormer~\cite{bevformer}   & C     & 41.6          & 51.7          \\
    \midrule
    CenterPoint~\cite{Yin2020Centerbased3O} & L     & 59.6          & 66.8          \\
    Focals Conv~\cite{focalsconv-chen} & L &        61.2    & 68.1  \\
    TransFusion-L~\cite{Bai2022TransFusionRL} & L & 65.1    & 70.1  \\
    \midrule
    TransFusion~\cite{Bai2022TransFusionRL} & LC    & 67.5 & 71.3 \\
    BEVFusion~\cite{Liang2022BEVFusionAS} & LC & 67.9   & 71.0 \\
    BEVFusion~\cite{Liu2022BEVFusionMM} & LC &  68.5    & 71.4\\
    CMT~\cite{Yan2023CrossMT}           & LC & 69.4 & 71.9 \\
    DeepInteraction~\cite{Yang2022DeepInteraction3O}      & LC & 69.9   & 72.6 \\
    \textbf{SimpleBEV}(ours) & LC                 & 71.4     & 73.5 \\
    \bottomrule
  \end{tabular}
  }
\caption{Performance comparison on the nuScenes validation set. No TTA and model ensemble are used.}
\label{table:val_comparison}
	\end{center}
\end{table}

\subsection{Ablation experiments}

\noindent\textbf{Ablation results about LiDAR branch}.
We evaluate the performance of the detector with only LiDAR data. The results are shown in Tab.~\ref{table:lidar_branch}. Compared with the CenterHead-based detector, the detector with the transformer-based head shows better performance. Meanwhile, we find that disabling the GT-paste augmentation strategy~\cite{second} for the last 5 epochs can significantly improve the detection performance, which called the fade strategy in \cite{Wang2021PointAugmentingCA}. As shown in the last two rows, fusing the multi-scaled BEV features and increasing the channel number can improve the detection performance.

\begin{table}
  \begin{center}
    \scalebox{0.95}{
        \begin{tabular}{ccccc|cc}
            \toprule
            C.H.            &   T.F.      &       Fad.        &    M.S.       &       $\times 2$ C.       & mAP       &   NDS \\
            \midrule
            $\checkmark$    &               &                   &               &                           & 54.08     &   62.59  \\
            $\checkmark$    &               &     $\checkmark$  &               &                           & 59.17     &   65.38  \\
                            & $\checkmark$  &                   &               &                           & 54.70     &   62.93  \\
                            & $\checkmark$  &     $\checkmark$  &               &                           & 60.63     &   66.32  \\
                            & $\checkmark$  &     $\checkmark$  & $\checkmark$  &                           & 61.71     &   66.93  \\
                            & $\checkmark$  &     $\checkmark$  & $\checkmark$  &  $\checkmark$             & 62.94     &   68.09  \\

            \bottomrule
        \end{tabular}
    }
    \caption{Ablating experiments of the LiDAR branch. `C.H.' denotes the CenterHead~\cite{Yin2020Centerbased3O}. `T.F.' represents the transformer-based head in \cite{Bai2022TransFusionRL}. `Fad.' represents the fade training strategy. `M.S.' denotes the multi-scaled feature fusion.}
    \label{table:lidar_branch}
  \end{center}
\end{table}

\noindent\textbf{Ablation of camera branch}.
We show the benefits of the improved camera branch in Tab.~\ref{table:camera_branch}. We first evaluate the performance of the detectors with different heads using only camera data. The CenterHead gains better performance. So, the auxiliary branch utilizes the CenterHead as the detection head. We first train a model with depth rectification by the LiDAR data. The results are shown in the fourth row. Then, we disable the depth rectification of the above model and show the performance in the third row. Without depth rectification, the detection performance drops by 2.08\% mAP and 1.22\% NDS. The depth information from the LiDAR data can improve the localization precision of image features and further bring better fusion detection performance. Meanwhile, the auxiliary branch can improve the final detection performance.

\begin{table}
  \begin{center}
    \scalebox{0.95}{
        \begin{tabular}{c|cccc|cc}
            \toprule
            Mod.    &    C.H.           &   T.F.      &   D.R.       &   Aux.        & mAP       &    NDS \\
            \midrule
                C   &    $\checkmark$   &               &               &               & 33.14     &   38.48   \\
                C   &                   & $\checkmark$  &               &               & 29.54     &   33.40   \\
            \midrule
                LC  &                   &  $\checkmark$ &               &               & 66.84     &  69.86    \\
                LC  &                   &  $\checkmark$ &   $\checkmark$&               & 68.92     &  71.08      \\
                LC  & $\checkmark$      & $\checkmark$  & $\checkmark$  & $\checkmark$  & 69.78     &  71.47     \\
            \bottomrule
        \end{tabular}
    }
    \caption{Ablating experiments of the camera branch. `C.H.' denotes the CenterHead~\cite{Yin2020Centerbased3O}. `T.F.' represents the transformer-based head in \cite{Bai2022TransFusionRL}. 'D.R.' denotes the depth rectification by the LiDAR data.}
    \label{table:camera_branch}
  \end{center}
\end{table}

\begin{table}
  \begin{center}
    \scalebox{0.95}{
        \begin{tabular}{cc|cc}
            \toprule
            img resolution                            &   grid size(m)            & mAP           & NDS        \\
            \midrule
            256$\times$704                      &   0.6             & 69.78         & 71.47      \\
            1600$\times$640                     &   0.6             & 70.96         & 73.22      \\
            256$\times$704                      &   0.3             & 71.35         & 73.47      \\
            \bottomrule
        \end{tabular}
    }
    \caption{Results of the detector with different image resolutions and BEV grid sizes.}
    \label{table:voxel_bev_size_details}
  \end{center}
\end{table}

We also evaluate the detectors with different image resolutions and BEV grid sizes on the nuScenes validation set. The results are shown in Tab.~\ref{table:voxel_bev_size_details}. Obviously, increasing the resolution of the image and BEV feature map can improve the 3D object detection performance.

\section{Conclusion}
In this paper, we propose an effective multi-modal fusion framework, SimpleBEV, to detect 3D objects in the autonomous driving environment. It follows the architecture of the BEV-based fusion method that fuses the LiDAR and camera features in a unified BEV space. The experiments demonstrate the effectiveness of our method. The improved camera depth estimation module and the multi-scaled LiDAR-BEV fusion module can effectively enhance the detection performance. Moreover, the introduced auxiliary branch benefits the camera information exploitation during training. In the future, we will integrate more sensors into the framework and explore more downstream applications based on the fused features.

{\small
\bibliographystyle{ieee_fullname}
\bibliography{SimpleBEV}
}

\end{document}